\documentclass[runningheads]{llncs}
\usepackage{graphicx}

\usepackage{tikz}
\usepackage{comment} 
\usepackage{amsmath,amssymb} 
\usepackage{color}

\usepackage{wrapfig,adjustbox}
\usepackage{enumitem}
\usepackage{booktabs}
\usepackage{subfigure}
\usepackage{multirow}
\usepackage{chngcntr}
\usepackage{xfrac}
\usepackage{array}
\usepackage{etoolbox}
\BeforeBeginEnvironment{wrapfigure}{\setlength{\intextsep}{0pt}}

\newcommand{\beginsupplement}{%
        \setcounter{table}{0}
        \renewcommand{\thetable}{S\arabic{table}}%
        \setcounter{figure}{0}
        \renewcommand{\thefigure}{S\arabic{figure}}%
        
        \setcounter{section}{0}
        \renewcommand{\thesection}{S\arabic{section}}%
        
        \setcounter{page}{1}
        \renewcommand{\thepage}{S\arabic{page}}%
}

\begin{document}
\pagestyle{headings}
\mainmatter
\def\ECCVSubNumber{100}  

\title{REMIND Your Neural Network to Prevent Catastrophic Forgetting} 

\titlerunning{REMIND Your Neural Network to Prevent Catastrophic Forgetting}
%
\author{Tyler L. Hayes\inst{1}$^{,\star}$ \and
Kushal Kafle\inst{2}$^{,\star}$ \and
Robik Shrestha\inst{1}$^,$\thanks{Equal Contribution.} \and \\
Manoj Acharya\inst{1} \and
Christopher Kanan\inst{1,3,4}
}
\authorrunning{Hayes et al.}
%
\institute{Rochester Institute of Technology, Rochester NY 14623, USA\\ \email{\{tlh6792,rss9369,ma7583,kanan\}@rit.edu} \and
Adobe Research, San Jose CA 95110, USA\\ \email{kkafle@adobe.com} \and
Paige, New York NY 10036, USA \and
Cornell Tech, New York NY 10044, USA
}
\maketitle

\begin{abstract}
People learn throughout life. However, incrementally updating conventional neural networks leads to catastrophic forgetting. A common remedy is replay, which is inspired by how the brain consolidates memory. Replay involves fine-tuning a network on a mixture of new and old instances. While there is neuroscientific evidence that the brain replays compressed memories, existing methods for convolutional networks replay raw images. Here, we propose REMIND, a brain-inspired approach that enables efficient replay with compressed representations. REMIND is trained in an online manner, meaning it learns one example at a time, which is closer to how humans learn. Under the same constraints, REMIND outperforms other methods for incremental class learning on the ImageNet ILSVRC-2012 dataset. We probe REMIND's robustness to data ordering schemes known to induce catastrophic forgetting. We demonstrate REMIND's generality by pioneering online learning for Visual Question Answering (VQA)\footnote{\url{https://github.com/tyler-hayes/REMIND}}. 
\end{abstract}

\section{Introduction}
\label{sec:intro}

The mammalian brain engages in continuous online learning of new skills, objects, threats, and environments. The world provides the brain a temporally structured stream of inputs, which is not independent and identically distributed (iid). Enabling online learning in artificial neural networks from non-iid data is known as lifelong learning. While conventional networks suffer from catastrophic forgetting~\cite{abraham2005memory,mccloskey1989}, with new learning overwriting existing representations, a wide variety of methods have recently been explored for overcoming this problem~\cite{castro2018end,chaudhry2019efficient,hou2019unified,kirkpatrick2017,lopez2017gradient,nguyen2018variational,rebuffi2016icarl,wu2019large}. Some of the most successful methods for mitigating catastrophic forgetting use variants of replay~\cite{castro2018end,hayes2019memory,hou2019unified,kemker2018forgetting,rebuffi2016icarl,wu2019large}, which involves mixing new instances with old ones and fine-tuning the network with this mixture. Replay is motivated by how the brain works: new experiences are encoded in the hippocampus and then these compressed memories are re-activated along with other memories so that the neocortex can learn them~\cite{lewis2011overlapping,o2010play,stickgold2001sleep}. Without the hippocampus, people lose the ability to learn new semantic categories~\cite{konkel2008hippocampal}. Replay occurs both during sleep~\cite{ji2007coordinated} and when awake~\cite{karlsson2009awake,takahashi2015episodic}.

For lifelong learning in convolutional neural networks (CNNs), there are two major gaps between existing methods and how animals learn. The first is that replay is implemented by storing and replaying raw pixels, which is not biologically plausible. Based on hippocampal indexing theory~\cite{teyler2007hippocampal},  the hippocampus stores \emph{compressed} representations of neocortical activity patterns while awake. To consolidate memories, these patterns are replayed and then the corresponding neocortical neurons are re-activated via reciprocal connectivity~\cite{lewis2011overlapping,o2010play,stickgold2001sleep}. The representations stored in the hippocampus for replay are not veridical (e.g., raw pixels)~\cite{ji2007coordinated,mcclelland1996considerations}, and its visual inputs are high in the visual processing hierarchy~\cite{insausti2017nonhuman} rather than from primary visual cortex or retina.

\begin{wrapfigure}[15]{r}{0.4\textwidth}
 \centering
      \includegraphics[width=\linewidth]{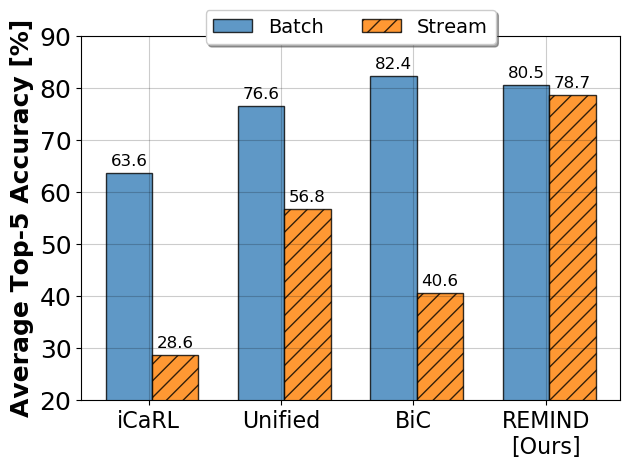}
  \caption{Average top-5 accuracy results for streaming and incremental batch versions of state-of-the-art models on ImageNet.  \label{fig:imagenet-stream-vs-batch}
  }
\end{wrapfigure}

The second major gap with existing approaches is that animals engage in \emph{streaming learning}~\cite{gama2010knowledge,gama2013evaluating}, or resource constrained online learning from non-iid (temporally correlated) experiences throughout life. In contrast, the most common paradigm for incremental training of CNNs is to break the training dataset into $M$ distinct batches, where for ImageNet each batch typically has about 100000 instances from 100 classes that are not seen in later batches, and then the algorithm sequentially loops over each batch many times. This paradigm is not biologically plausible. There are many applications requiring online learning of non-iid data streams, where batched learning will not suffice, such as immediate on-device learning. Batched systems also take longer to train, further limiting their utility on resource constrained devices, such as smart appliances, robots, and toys. For example, BiC, a state-of-the-art incremental batch method, requires 65 hours to train in that paradigm whereas our proposed streaming model trains in under 12 hours. The incremental batch setting can be transformed into the streaming learning scenario by using very small batches and performing only a single pass through the dataset; however, this results in a large decrease in performance. As shown in Fig.~\ref{fig:imagenet-stream-vs-batch}, state-of-the-art methods perform poorly on ImageNet in the streaming setting, with the best method suffering an over 19\% drop in performance. In contrast, our model outperforms the best streaming model by 21.9\% and is only 1.9\% below the best batch model.

Here, we propose REMIND, or \textbf{re}play using \textbf{m}emory \textbf{ind}exing, a novel method that is heavily influenced by biological replay and hippocampal indexing theory. \textbf{Our main contributions are:}
\begin{enumerate}[noitemsep, nolistsep]

\item We introduce REMIND, a streaming learning model that implements hippocampal indexing theory using tensor quantization to efficiently store hidden representations (e.g., CNN feature maps) for later replay. REMIND implements this compression using Product Quantization (PQ)~\cite{jegou2010product}. We are the first to test if forgetting in CNNs can be mitigated by replaying hidden representations rather than raw pixels.

\item REMIND outperforms existing models on the ImageNet ILSVRC-2012~\cite{russakovsky2015imagenet} and CORe50~\cite{lomonaco2017core50} datasets, while using the same amount of memory.

\item  We demonstrate REMIND's robustness by pioneering streaming Visual Question Answering (VQA), in which an agent must answer questions about images and cannot be readily done with existing models. We establish new experimental paradigms, baselines, and metrics and subsequently achieve strong results on the CLEVR~\cite{johnson2017clevr} and TDIUC~\cite{kafle2017analysis} datasets.

\end{enumerate}

\section{Problem Formulation}
\label{sec:problem-formulation}

There are multiple paradigms in which incremental learning has been studied~\cite{parisi2019continual}. In \emph{incremental batch learning}, at each time step $t$ an agent learns a data batch $B_{t}$ containing $N_{t}$ instances and their corresponding labels, where $N_{t}$ is often 1000 to 100000. While much recent work has focused on incremental batch learning~\cite{castro2018end,chaudhry2018riemannian,fernando2017,hou2019unified,kemker2018fearnet,kemker2018forgetting,rebuffi2016icarl,wu2019large,zenke2017continual}, \emph{streaming learning}, or online learning from non-iid data streams with memory and/or compute constraints, more closely resembles animal learning and has many applications~\cite{gama2010knowledge,gama2013evaluating,le2017expressiveness}. In streaming learning, a model learns online in a single pass, i.e., $N_{t}=1$ for all $t$. It cannot loop over any portion of the (possibly infinite) dataset, and it can be evaluated at any point rather than only between large batches. Streaming learning can be approximated by having a system queue up small, temporally contiguous, mini-batches for learning, but as shown in Fig.~\ref{fig:imagenet-stream-vs-batch}, batch methods cannot easily adapt to this setting.

\section{Related Work}
\label{sec:related work}

Parisi et al.~\cite{parisi2019continual} identify three main mechanisms for mitigating forgetting in neural networks, namely 1) replay of previous knowledge, 2) regularization mechanisms to constrain parameter updates, and 3) expanding the network as more data becomes available. Replay has been shown to be one of the most effective methods for mitigating catastrophic forgetting~\cite{aljundi2019online,aljundi2019gradient,castro2018end,hayes2019memory,hou2019unified,kemker2018fearnet,kemker2018forgetting,lee2019overcoming,ostapenko2019learning,rebuffi2016icarl,wu2019large}. For ImageNet, all recent state-of-the-art methods for incremental class learning use replay of raw pixels with distillation loss. The earliest was iCaRL~\cite{rebuffi2016icarl}, which stored 20 images per class for replay. iCaRL used a nearest class prototype classifier to mitigate forgetting. The End-to-End incremental learning model~\cite{castro2018end} extended iCaRL to use the outputs of the CNN directly for classification, instead of a nearest class mean classifier. Additionally, End-to-End used more data augmentation and a balanced fine-tuning stage during training to improve performance. The Unified classifier~\cite{hou2019unified} extended End-to-End by using a cosine normalization layer, a new loss constraint, and a margin ranking loss. The Bias Correction (BiC)~\cite{wu2019large} method extended End-to-End by training two additional parameters to correct bias in the output layer due to class imbalance. iCaRL, End-to-End, the Unified classifier, and BiC all: 1) store the same number of raw replay images per class, 2) use the same herding procedure for prototype selection, and 3) use distillation loss to prevent forgetting. REMIND, however, is the first model to demonstrate that storing and replaying quantized mid-level CNN features is an effective strategy to mitigate forgetting.

Regularization methods vary a weight's plasticity based on how important it is to previous tasks. These methods include Elastic Weight Consolidation (EWC)~\cite{kirkpatrick2017}, Memory Aware Synapses (MAS)~\cite{aljundi2018memory}, Synaptic Intelligence (SI)~\cite{zenke2017continual}, Riemannian Walk (RWALK)~\cite{chaudhry2018riemannian}, Online Laplace Approximator~\cite{ritter2018online}, Hard Attention to the Task~\cite{serra2018overcoming}, and Learning without Memorizing~\cite{dhar2019learning}. The Averaged Gradient Episodic Memory (A-GEM)~\cite{chaudhry2019efficient} model extends Gradient Episodic Memory~\cite{lopez2017gradient}, which uses replay with regularization. Variational Continual Learning~\cite{nguyen2018variational} combines Bayesian inference with replay, while the Meta-Experience Replay model~\cite{riemer2018learning} combines replay with meta-learning. All of these regularization methods are typically used for incremental \emph{task} learning, where batches of data are labeled as different tasks and the model must be told which task (batch) a sample came from during inference. When task labels are not available at test time, which is often true for agents operating in real-time, many methods cannot be used or they will fail~\cite{chaudhry2018riemannian,farquhar2018towards,kemker2018forgetting}. While our main experiments focus on comparisons against state-of-the-art ImageNet models, we compare REMIND against several regularization models in Sec.~\ref{sec: additional-experiments}, both with and without task labels. Some regularization methods also utilize cached data, e.g., GEM and A-GEM.

Another approach to mitigating forgetting is to expand the network as new tasks are observed, e.g., Progressive Neural Networks~\cite{rusu2016progressive}, Dynamically Expandable Networks~\cite{yoon2018lifelong}, Adaptation by Distillation~\cite{hou2018lifelong}, and Dynamic Generative Memory~\cite{ostapenko2019learning}. However, these approaches also use task labels at test time, have growing memory requirements, and may not scale to thousand-category datasets.

\section{REMIND: Replay using Memory Indexing}
\label{sec:model}

\begin{figure}[t]
 \centering
    \includegraphics[width=0.75\linewidth]{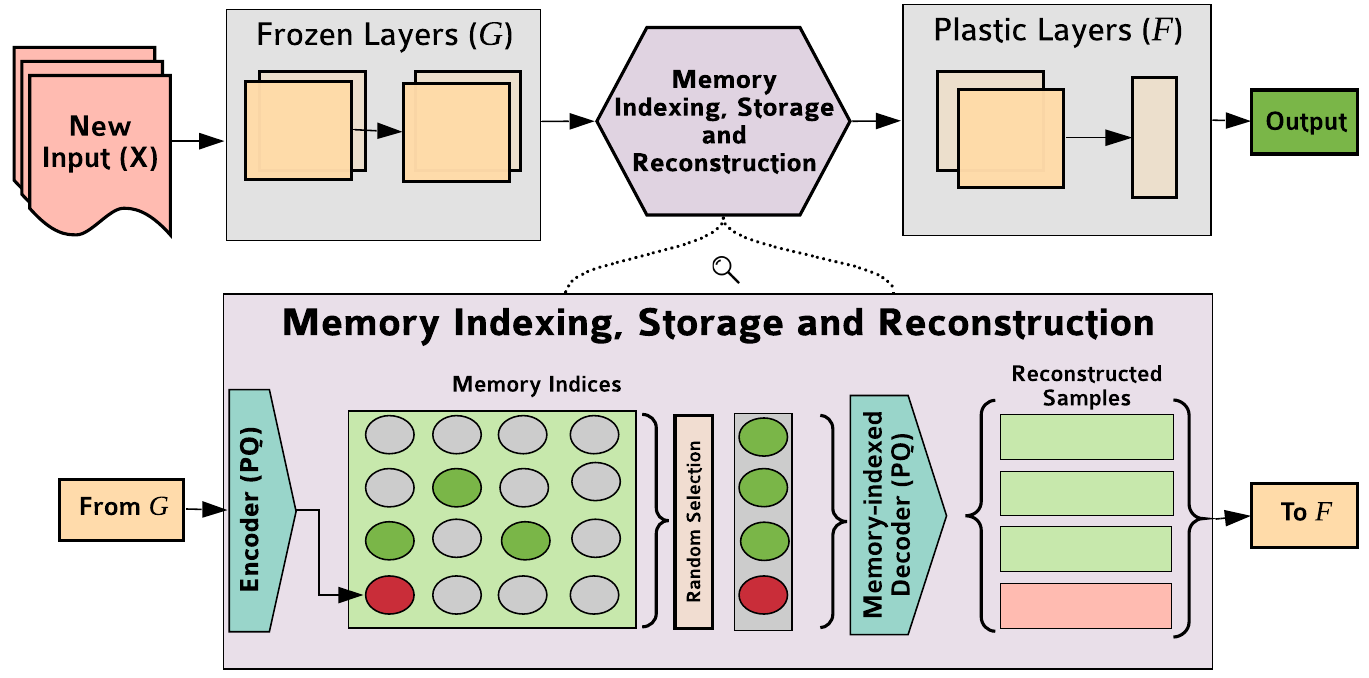}
      \caption{REMIND takes in an input image and passes it through frozen layers of the network ($G$) to obtain tensor representations (feature maps). It then quantizes the tensors via product quantization and stores the indices in memory for future replay. The decoder reconstructs tensors from the stored indices to train the plastic layers ($F$) of the network before a final prediction is made.}
      \label{fig: remind-model}
\end{figure}

REMIND is a novel brain-inspired method for training the parameters of a CNN in the streaming setting using replay. Learning involves two steps: 1) compressing the current input and 2) reconstructing a subset of previously compressed representations, mixing them with the current input, and updating the \emph{plastic} weights of the network with this mixture (see Fig.~\ref{fig: remind-model}). While earlier work for incremental batch learning with CNNs stored raw images for replay~\cite{castro2018end,hou2019unified,rebuffi2016icarl,wu2019large}, by storing compressed mid-level CNN features, REMIND is able to store far more instances with a smaller memory budget. For example, iCaRL~\cite{rebuffi2016icarl} uses a default memory budget of 20K examples for ImageNet, but REMIND can store over 1M compressed instances using the same budget. This more closely resembles how replay occurs in the brain, with high-level visual representations being sent to the hippocampus for storage and re-activation, rather than early visual representations~\cite{insausti2017nonhuman}. REMIND does not have an explicit sleep phase, with replay more closely resembling that during waking hours~\cite{karlsson2009awake,takahashi2015episodic}.

Formally, our CNN $y_i = F\left( G \left( \mathbf{X}_i \right) \right)$ is trained in a streaming paradigm, where $\mathbf{X}_i$ is the input image and $y_i$ is the predicted output category. The network is composed of two nested functions: $G\left( \cdot \right)$, parameterized by $\theta_G$, consists of the first $J$ layers of the CNN and $F\left( \cdot \right)$, parameterized by $\theta_F$, consists of the last $L$ layers. REMIND keeps $\theta_G$ fixed since early layers of CNNs have been shown to be highly transferable~\cite{yosinski2014transferable}. The later layers, $F\left( \cdot \right)$, are trained in the streaming paradigm using REMIND. We discuss how $G\left( \cdot \right)$ is initialized in Sec.~\ref{sec:base-initialization}.

The output of $G\left( \mathbf{X}_i \right)$ is a tensor $\mathbf{Z}_i \in \mathbb{R}^{m \times m \times d}$, where $m$ is the dimension of the feature map and $d$ is the number of channels. Using the outputs of $G\left(\cdot\right)$, we train a vector quantization model for the $\mathbf{Z}_i$ tensors. As training examples are observed, the quantization model is used to store the $\mathbf{Z}_i$ features and their labels in a replay buffer as an $m \times m \times s$ array of integers using as few bits as necessary, where $s$ is the number of indices that will be stored. For replay, we uniformly select $r$ instances from the replay buffer, which was shown to work well in~\cite{chaudhry2018riemannian}, and reconstruct them. Each of the reconstructed instances, $\hat{\mathbf{Z}}_{i}$, are mixed with the current input, and then $\theta_F$ is updated using backpropagation on this set of $r+1$ instances. Other selection strategies are discussed in Sec.~\ref{sec:discussion}. During inference, we pass an image through $G\left( \cdot \right)$, and then the output, $\mathbf{Z}_i$, is quantized and reconstructed before being passed to $F\left( \cdot \right)$.

Our main version of REMIND uses PQ~\cite{jegou2010product} to compress and store $\mathbf{Z}_i$. For high-dimensional data, PQ tends to have much lower reconstruction error than models that use only $k$-means. The tensor $\mathbf{Z}_i$ consists of $m \times m$ $d$-dimensional tensor elements, and PQ partitions each $d$-dimensional tensor element into $s$ sub-vectors, each of size $d / s$. PQ then creates a separate codebook for each partition by using $k$-means, where the codes within each codebook correspond to the centroids learned for that partition. Since the quantization is done independently for each partition, each sub-vector of the $d$-dimensional tensor element is assigned a separate integer, so the element is represented with $s$ integers. If $s$ is equal to one, then this approach is identical to using $k$-means for vector quantization, which we compare against. For our experiments, we set $s=32$ and $c=256$, so that each integer can be stored with 1 byte. We explore alternative values of $s$ and $c$ in supplemental materials (Fig.~\ref{fig:additional-experiments-cls}) and use the Faiss PQ implementation~\cite{FAISS}.

Since lifelong learning systems must be capable of learning from infinitely long data streams, we subject REMIND's replay buffer to a maximum memory restriction. That is, REMIND stores quantization indices in its buffer until this maximum capacity has been reached. Once the buffer is full and a new example comes in, we insert the new sample and randomly remove an example from the class with the most examples, which was shown to work well in~\cite{chaudhry2018riemannian,wu2019large}. We discuss other strategies for maintaining the replay buffer in Sec.~\ref{sec:discussion}.

\subsection{Augmentation During Replay}
\label{sec:memory-replay}

To augment data during replay, REMIND uses random resized crops and a variant of manifold mixup~\cite{verma2018manifold} on the quantized tensors directly. For random crop augmentation, the tensors are randomly resized, then cropped and bilinearly interpolated to match the original tensor dimensions. To produce more robust representations, REMIND mixes features from multiple classes using manifold mixup. That is, REMIND uses its replay buffer to reconstruct two randomly chosen sets, $\mathcal{A}$ and $\mathcal{B}$, of $r$ instances each ($\lvert \mathcal{A}\rvert=\lvert \mathcal{B}\rvert=r$), which are linearly combined to obtain a set $\mathcal{C}$ of $r$ mixed instances ($\lvert \mathcal{C}\rvert=r$), i.e., a newly mixed instance, $\left( \mathbf{Z}_{\mathrm{mix}}, y_{\mathrm{mix}} \right) \in \mathcal{C}$, is formed as:
\begin{equation}
\left( \mathbf{Z}_{\mathrm{mix}}, y_{\mathrm{mix}} \right) = \left(\lambda \mathbf{Z}_{a} + \left(1 - \lambda \right) \mathbf{Z}_{b}, \lambda y_{a} + \left(1 - \lambda \right)y_{b}\right),
\end{equation}
where $\left(\mathbf{Z}_{a},y_a\right)$ and $\left(\mathbf{Z}_{b},y_b\right)$ denote instances from $\mathcal{A}$ and $\mathcal{B}$ respectively and $\lambda \sim \beta(\alpha, \alpha)$ is the mixing coefficient drawn from a $\beta$-distribution parameterized by hyperparameter $\alpha$. We use $\alpha=0.1$, which we found to work best in preliminary experiments. The current input is then combined with the set $\mathcal{C}$ of $r$ mixed samples, and $\theta_F$ is updated using this new set of $r+1$ instances.

\subsection{Initializing REMIND}
\label{sec:base-initialization}

During learning, REMIND only updates $F\left( \cdot \right)$, i.e., the top of the CNN. It assumes that $G\left( \cdot \right)$, the lower level features of the CNN, are fixed. This implies that the low-level visual representations must be highly transferable across image datasets, which is supported empirically~\cite{yosinski2014transferable}. There are multiple methods for training $G\left( \cdot \right)$, including supervised pre-training on a portion of the dataset, supervised pre-training on a different dataset, or unsupervised self-taught learning using a convolutional auto-encoder. Here, we follow the common practice of doing a `base initialization' of the CNN~\cite{castro2018end,hou2019unified,rebuffi2016icarl,wu2019large}. This is done by training both $\theta_F$ and $\theta_G$ jointly on an initial subset of data offline, e.g., for class incremental learning on ImageNet we use the first 100 classes. After base initialization, $\theta_G$ is no longer plastic. All of the examples $\mathbf{X}_i$ in the base initialization are pushed through the model to obtain $\mathbf{Z}_i = G\left( \mathbf{X}_i \right)$, and all of these $\mathbf{Z}_i$ instances are used to learn the quantization model for $G\left( \mathbf{X}_i \right)$, which is kept fixed once acquired.

Following \cite{castro2018end,hou2019unified,rebuffi2016icarl,wu2019large}, we use ResNet-18~\cite{He_2016_CVPR} for image classification, where we set $G\left( \cdot \right)$ to be the first 15 convolutional and 3 downsampling layers, which have 6,455,872 parameters, and $F\left( \cdot \right)$ to be the remaining 3 layers (2 convolutional and 1 fully connected), which have 5,233,640 parameters. These layers were chosen for memory efficiency in the quantization model with ResNet-18, and we show the memory efficiency trade-off in supplemental materials (Fig.~\ref{fig:resnet-memory}).

\section{Experiments: Image Classification}
\label{sec:image-classification}

\subsection{Comparison Models}
While REMIND learns on a per sample basis, most methods for incremental learning in CNNs do multiple loops through a batch. For fair comparison, we train these methods in the streaming setting to fairly compare against REMIND. Results for the incremental batch setting for these models are included in Fig.~\ref{fig:imagenet-stream-vs-batch} and supplemental materials (Table~\ref{tab:averaged-results} and Fig.~\ref{fig:imagenet-batch-memory}-\ref{fig:core50-memory}). We evaluate the following:
\begin{itemize}[noitemsep, nolistsep]
    \item \textbf{REMIND} -- Our main REMIND version uses PQ and replay augmentation. We also explore a version that omits data augmentation and a version that uses $k$-means rather than PQ.
    \item \textbf{Fine-Tuning (No Buffer)} -- Fine-Tuning is a baseline that fine-tunes $\theta_F$ of a CNN one sample at a time with a single epoch through the dataset. This approach does not use a buffer and suffers from catastrophic forgetting~\cite{kemker2018forgetting}. 
    \item \textbf{ExStream} -- Like REMIND, ExStream is a streaming learning method, however, it can only train fully connected layers of the network~\cite{hayes2019memory}. ExStream uses rehearsal by maintaining buffers of prototypes. It stores the input vector and combines the two nearest vectors in the buffer. After the buffer gets updated, all samples from its buffer are used to train the fully connected layers of a network. We use ExStream to train the final layer of the network, which is the only fully connected layer in ResNet-18.
    \item \textbf{SLDA} -- Streaming Linear Discriminant Analysis (SLDA) is a well-known streaming method that was shown to work well on deep CNN features~\cite{hayes2019lifelong}. It maintains running means for each class and a running tied covariance matrix. Given a new input, it assigns the label of the closest Gaussian in feature space. It can be used to compute the output layer of a CNN.
    \item \textbf{iCaRL} -- iCaRL is an incremental batch learning algorithm for CNNs~\cite{rebuffi2016icarl}. iCaRL stores images from earlier classes for replay, uses a distillation loss to preserve weights, and uses a nearest class mean classifier in feature space. 
    \item \textbf{Unified} -- The Unified Classifier builds on iCaRL by using the outputs from the network for classification and introducing a cosine normalization layer, a constraint to preserve class geometry, and a margin ranking loss to maximize inter-class separation~\cite{hou2019unified}. Unified also uses replay and distillation.
    \item \textbf{BiC} -- The Bias Correction (BiC) method builds on iCaRL by using the output layer of the network for classification and correcting the bias from class imbalance during training, i.e., more new samples than replay samples~\cite{wu2019large}. The method trains two additional bias correction parameters on the output layer, resulting in improved performance over distillation and replay alone.
    \item \textbf{Offline} -- The offline model is trained in a traditional, non-streaming setting and serves as an upper-bound on performance. We train two variants: one with only $\theta_F$ plastic and one with both $\theta_F$ and $\theta_G$ plastic.
\end{itemize}
Our main experiments focus on comparing state-of-the-art methods on ImageNet and we provide additional comparisons in Sec.~\ref{sec: additional-experiments}. Although iCaRL, Unified, and BiC are traditionally trained in the incremental batch paradigm, we conduct experiments with these models in the streaming paradigm for fair comparison against REMIND. To train these streaming variants, we set the number of epochs to 1 and the batch size to $r+1$ instances to match REMIND.

\subsection{Model Configurations}

In our setup, all models are trained instance-by-instance and have no batch requirements, unless otherwise noted. Because methods can be sensitive to the order in which new data are encountered, all models receive examples in the same order. The same base CNN initialization procedure is used by all models. For ExStream and SLDA, after base initialization, the streaming learning phase is re-started from the beginning of the data stream. All of the parameters except the output layer are kept frozen for ExStream and SLDA, whereas only $G\left( \cdot \right)$ is kept frozen for REMIND. All other comparison models do not freeze any layers and incremental training commences with the first new data sample. All models, except SLDA, are trained using cross-entropy loss with stochastic gradient descent and momentum. More parameter settings are in supplemental materials.

\subsection{Datasets, Data Orderings, \& Metrics}
\label{sec:ic-datasets}

We conduct experiments with ImageNet and CORe50 by dividing both datasets into batches. The first batch is used for base initialization. Subsequently, all models use the same batch orderings, but they are sequentially fed individual samples and they cannot revisit any instances in a batch, unless otherwise noted. For ImageNet, the models are evaluated after each batch on all trained classes. For CORe50, models are evaluated on all test data after each batch.

ImageNet ILSVRC-2012~\cite{russakovsky2015imagenet} has 1000 categories each with 732-1300 training samples and 50 validation samples, which we use for testing. During the base initialization phase, the model is trained offline on a set of 100 randomly selected classes. Following \cite{castro2018end,hou2019unified,rebuffi2016icarl,wu2019large}, each incremental batch then contains 100 random classes, which are not contained within any other batch. We study class incremental (class iid) learning with ImageNet.

CORe50~\cite{lomonaco2017core50} contains sequences of video frames, with one object in each frame. It has 10 classes, and each sequence is acquired with varied environmental conditions. CORe50 is ideal for evaluating streaming learners since it is naturally non-iid and requires agents to learn from temporally correlated video streams. For CORe50, we follow \cite{hayes2019memory} and sample at 1 frame per second, obtaining 600 training images and 225 test images per class.  We use the bounding box crops and splits from \cite{lomonaco2017core50}. Following \cite{hayes2019memory}, we use four training orderings to test the robustness of each algorithm under different conditions: 1) iid, where each batch has a random subset of training images, 2) class iid, where each batch has all of the images from two classes, which are randomly shuffled, 3) instance, where each batch has temporally ordered images from 80 unique object instances, and 4) class instance, where each batch has all of the temporally ordered instances from two classes. All batches have 1200 images across all orderings. Since CORe50 is small, CNNs are first initialized with pre-trained ImageNet weights and then fine-tuned on a subset of 1200 samples for base initialization.

We use the $\Omega_{\mathrm{all}}$ metric~\cite{hayes2019memory,hayes2018new,kemker2018forgetting} for evaluation, which normalizes incremental learning performance by offline performance:
$
\Omega_{\mathrm{all}} = \frac{1}{T} \sum_{t=1}^T \frac{\alpha_{t}}{\alpha_{\mathrm{offline},t}} \enspace ,
$
where $T$ is the total number of testing events, $\alpha_{t}$ is the accuracy of the model for test $t$, and $\alpha_{\mathrm{offline},t}$ is the accuracy of the optimized offline learner for test $t$. If $\Omega_{\mathrm{all}} = 1$, then the incremental learner's performance matched the offline model. We use top-5 and top-1 accuracies for ImageNet and CORe50, respectively. Average accuracy results are in supplemental materials (Table~\ref{tab:averaged-results}-\ref{tab:averaged-results-core50}).

\subsection{Results: ImageNet}
\label{sec:ic results}

\begin{wrapfigure}[13]{r}{0.4\textwidth} 
\centering
      \includegraphics[width=\linewidth]{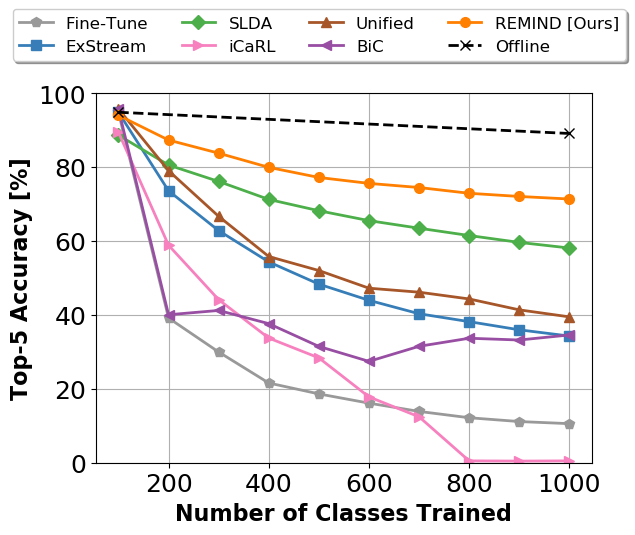}
  \caption{Performance of streaming ImageNet models. \label{fig:forgetting-curve} 
  }
\end{wrapfigure}

For ImageNet, we use the pre-trained PyTorch offline model with 89.08\% top-5 accuracy to normalize $\Omega_{\mathrm{all}}$. We allow the iCaRL, Unified, and BiC models to store 10,000 (224$\times$224 uint8) raw pixel image prototypes in a replay buffer, which is equivalent to 1.51 GB in memory. This allows REMIND to store indices for 959665 examples in its replay buffer. We set $r=50$ samples. We study additional buffer sizes in Sec.~\ref{sec: additional-experiments}. Results for incremental class learning on ImageNet are shown in Table~\ref{tab:classification-results} and a learning curve for all models is shown in Fig.~\ref{fig:forgetting-curve}. REMIND outperforms all other comparison models, with SLDA achieving the second best performance. This is remarkable since REMIND only updates $\theta_F$, whereas iCaRL, Unified, and BiC all update $\theta_F$ and $\theta_G$. 

REMIND is intended to be used for online streaming learning; however, we also created a variant suitable for incremental batch learning which is described in supplemental materials. Incremental batch results for REMIND and recent methods are given in Fig.~\ref{fig:imagenet-stream-vs-batch} and supplemental materials (Table~\ref{tab:averaged-results} and Fig.~\ref{fig:imagenet-batch-memory}). While incremental batch methods train much more slowly, REMIND achieves comparable performance to the best methods.

\begin{table*}[t]
\caption{ResNet-18 streaming classification results on ImageNet and CORe50 using $\Omega_{\mathrm{all}}$. For CORe50, we explore performance across four ordering schemes and report the average of 10 permutations. Upper bounds are at the bottom.
}
\label{tab:classification-results}
\centering
\begin{tabular}{lccccc}
\toprule
& \textsc{\textbf{ImageNet}} & \multicolumn{4}{c}{\textsc{\textbf{CORe50}}} \\ 
 \cmidrule(r){2-2} \cmidrule(r){3-6}
\textsc{Model} & \textsc{cls iid} & \textsc{iid} & \textsc{cls iid} & \textsc{inst} & \textsc{cls inst} \\ 
\midrule
Fine-Tune ($\theta_F$) & 0.288 & 0.961 & 0.334 & 0.851 & 0.334 \\
ExStream & 0.569 & 0.953 & 0.873 & 0.933 & 0.854 \\
SLDA  & 0.752 & 0.976 & 0.958 & 0.963 & 0.959 \\
iCaRL & 0.306 & - & 0.690 & - & 0.644 \\
Unified & 0.614 & - & 0.510 & - & 0.527 \\
BiC & 0.440 & - & 0.410 & - & 0.415 \\
REMIND & \textbf{0.855} & \textbf{0.985} & \textbf{0.978} & \textbf{0.980} & \textbf{0.979} \\ 
\midrule
Offline ($\theta_F$) & 0.929 & 0.989 & 0.984 & 0.985 & 0.985 \\
Offline  & 1.000 & 1.000 & 1.000 & 1.000 & 1.000 \\
\bottomrule
\end{tabular}
\end{table*}

\subsection{Results: CORe50}

We use the CoRe50 dataset to study models under more realistic data orderings. Existing methods including iCaRL, Unified, and BiC assume that classes from one batch do not appear in other batches, making it difficult for them to learn the iid and instance orderings without modifications. To compute $\Omega_{\mathrm{all}}$, we use an offline model that obtains 93.11\% top-1 accuracy. The iCaRL, Unified, and BiC models use replay budgets of 50 images, which is equivalent to 7.3 MB.  This allows REMIND to store replay indices for 4465 examples. Results for other buffer sizes are in supplemental materials (Fig.~\ref{fig:core50-memory}). REMIND replays $r=20$ samples. $\Omega_{\mathrm{all}}$ results for CORe50 are provided in Table~\ref{tab:classification-results}. For CORe50, REMIND outperforms all models for all orderings. In fact, REMIND is only 2.2\% below the full offline model in the worst case, in terms of $\Omega_{\mathrm{all}}$. Methods that only trained the output layer performed well on CORe50 and poorly on ImageNet.  This is likely because the CNNs used for CORe50 experiments are initialized with ImageNet weights, resulting in more robust representations. REMIND's remarkable performance on these various orderings demonstrate its versatility.

\section{Experiments: Incremental VQA}
\label{sec:streaming vqa}

In VQA, a system must produce an answer to a natural language question about an image~\cite{antol2015vqa,kafle2016review,malinowski2014multi}, which requires capabilities such as object detection, scene understanding, and logical reasoning. Here, we use REMIND to pioneer streaming VQA. During training, a streaming VQA model receives a sequence of temporally ordered triplets $\mathcal{D}=\left\{\left(X_{t}, Q_{t}, A_{t}\right)\right\}_{t=1}^{T}$, where $X_t$ is an image, $Q_t$ is the question (string), and $A_t$ is the answer. If an answer is not provided at time $t$, then the agent must use knowledge from time $1$ to $t-1$ to predict $A_{t}$. To use REMIND for streaming VQA, we store each quantized feature along with a question string and answer, which can later be used for replay. REMIND can be used with almost any existing VQA system (e.g., attention-based~\cite{Anderson2017up-down,Kim2018BilinearAN,Yang2016}, compositional~\cite{andreas2016neural,arad2018compositional}, bi-modal fusion~\cite{ben2017mutan,FukuiPYRDR16,shrestha2019ramen}) and it can be applied to similar tasks like image captioning~\cite{bernardi2016automatic} and referring expression 
recognition~\cite{kazemzadeh2014referitgame,plummer2015flickr30k,rohrbach2016grounding}.

\subsection{Experimental Setup}

For our experiments, we use the TDIUC~\cite{kafle2017analysis} and CLEVR~\cite{johnson2017clevr} VQA datasets. TDIUC is composed of natural images and has over 1.7 million QA pairs organized into 12 question types including simple object recognition, complex counting, positional reasoning, and attribute classification. TDIUC tests for generalization across different underlying tasks required for VQA. CLEVR consists of over 700000 QA pairs for 70000 synthetically generated images and is organized into 5 question types. CLEVR specifically tests for multi-step compositional reasoning that is very rarely encountered in natural image VQA datasets. We combine REMIND with two popular VQA algorithms, using a modified version of the stacked attention network (SAN)~\cite{Kazemi2017ShowAA,Yang2016} for TDIUC, and a simplified version of the Memory Attention and Control (MAC)~\cite{arad2018compositional,marois2018transfer} network for CLEVR. A ResNet-101 model pre-trained on ImageNet is used to extract features for both TDIUC and CLEVR. REMIND's PQ model is trained with 32 codebooks each of size 256. The final offline mean per-type accuracy with SAN on TDIUC is 67.59\% and the final offline accuracy with MAC on CLEVR is 94.00\%. Our main results with REMIND use a buffer consisting of 50\% of the dataset and $r=50$. Results for other buffer sizes are in supplemental materials (Table~\ref{tab:vqa-ablations}).

For both datasets, we explore two orderings of the training data: iid and question type (q-type). For iid, the dataset is randomly shuffled and the model is evaluated on all test data when multiples of 10\% of the total training set are seen. The q-type ordering reflects a more interesting scenario where QA pairs for different VQA `skills' are grouped together. Models are evaluated on all test data at the end of each q-type. We perform base initialization by training on the first 10\% of the data for the iid ordering and on QA pairs belonging to the first q-type for the q-type ordering. Then, the remaining data is streamed into the model one sample at a time. The buffer is then incrementally updated with PQ encoded features and raw question strings. We use simple accuracy for CLEVR and mean-per-type accuracy for TDIUC.

We compare REMIND to ExStream~\cite{hayes2019memory}, SLDA~\cite{hayes2019lifelong}, an offline baseline, and a simple baseline where models are fine-tuned without a buffer, which causes catastrophic forgetting. To adapt ExStream and SLDA for VQA, we use a variant of the linear VQA model in \cite{Kafle2016}, which concatenates ResNet-101 image features to question features extracted from a universal sentence encoder~\cite{Subramanian2018LearningGP} and then trains a linear classifier. Parameter settings are in supplemental materials.

\subsection{Results: VQA}
\label{sec:vqa results}

\begin{wraptable}[12]{r}{0.5\textwidth}
\caption{$\Omega_{\mathrm{all}}$ results for streaming VQA.  \label{tab:vqa-results}}
\centering
\resizebox{0.48\textwidth}{!}{%
\begin{tabular}{lcccc}
\toprule
& \multicolumn{2}{c}{\textsc{TDIUC}} & \multicolumn{2}{c}{\textsc{CLEVR}} \\
\cmidrule(r){2-3} \cmidrule(r){4-5}
\textsc{Ordering} & \textsc{iid} & \textsc{q-type} & \textsc{iid} & \textsc{q-type} \\ 
\midrule
Fine-Tune & 0.716 & 0.273 & 0.494 & 0.260 \\
ExStream & 0.676  & 0.701 & 0.477 & 0.375 \\
SLDA & 0.624 & 0.644 & 0.518 & 0.496 \\
REMIND & \textbf{0.917} & \textbf{0.919} & \textbf{0.720} & \textbf{0.985} \\
\midrule
Offline & 1.000 & 1.000 & 1.000 & 1.000 \\
\bottomrule
\end{tabular}
}
\end{wraptable}

Streaming VQA results for REMIND with a 50\% buffer size are given in Table~\ref{tab:vqa-results}. Variants of REMIND with other buffer sizes are in supplemental materials (Table~\ref{tab:vqa-ablations}). REMIND outperforms the streaming baselines for both datasets, with strong performance on both TDIUC using the SAN model and CLEVR using the MAC model. Interestingly, for CLEVR the results are much greater for q-type than for iid. We hypothesize that the q-type ordering may be acting as a natural curriculum~\cite{bengio_curriculum_2009}, allowing our streaming model to train more efficiently. Our results demonstrate that it is possible to train complex, multi-modal agents capable of attention and compositional reasoning in a streaming manner. Learning curves and qualitative examples are in supplemental materials (Fig.~\ref{fig:main-results-vqa}-\ref{fig:qualitative-vqa}).


\section{Additional Classification Experiments}
\label{sec: additional-experiments}

In this section, we study several of REMIND's components. In supplemental materials, we study other factors that influence REMIND's performance (Fig.~\ref{fig:additional-experiments-cls}), e.g., where to quantize, number of codebooks, codebook size, and replay samples ($r$). In supplemental materials, we also explore the performance of iCaRL, Unified, and BiC when only $\theta_F$ is updated (Sec.~\ref{sec:updating-only-F}).

\subsubsection{REMIND Components.}

\begin{wraptable}[11]{r}{0.5\textwidth}
\caption{REMIND variations on ImageNet with their memory (GB).
}
\label{tab:remind-ablations}
\centering
\begin{tabular}{lcc}
\toprule

\textsc{Variant} & $\Omega_{\mathrm{all}}$ & \textsc{Memory} \\
\midrule

REMIND (Main) & 0.855 & 1.51 \\ 
100\% Buffer & 0.856 & 2.01 \\ 
No Augmentation & 0.818 & 1.51 \\
$k$-Means & 0.778 & 0.12 \\
Real Features & 0.868 & 24.08 \\
\bottomrule
\end{tabular}
\end{wraptable}

REMIND is impacted by the size of its overall buffer, using augmentation, and the features used to train $F(\cdot)$. We study these on ImageNet and results are given in Table~\ref{tab:remind-ablations}. REMIND (Main) denotes the variant of REMIND from our main experiments that uses augmentation with a buffer size of 959665 and 32 codebooks of size 256. PQ is critical to performance, with PQ (32 codebooks) outperforming $k$-means (1 codebook) by 7.7\% in terms of $\Omega_{\mathrm{all}}$. Augmentation is the next most helpful component and improves performance by 3.7\%. Storing the entire dataset (100\% Buffer) does not yield significant improvements. Using real features yields marginal improvements (1.3\%) while requiring nearly 16 times more memory.

\subsubsection{Replay Buffer Size.}

\begin{wrapfigure}[14]{r}{0.4\textwidth}
 \centering
      \includegraphics[width=\linewidth]{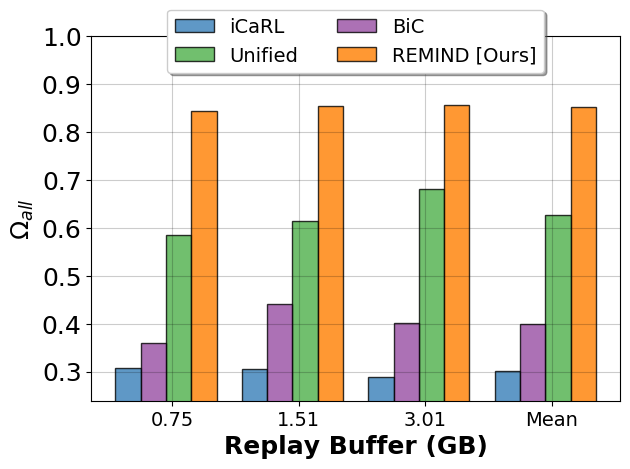}
		\caption{$\Omega_{\mathrm{all}}$ as a function of buffer size for streaming ImageNet models.}
		\label{fig:imagenet-stream-memory}
\end{wrapfigure}

Since REMIND and several other models rely on a replay buffer to mitigate forgetting, we studied performance on ImageNet as a function of buffer size. We compared the performance of iCaRL, Unified, and BiC on ImageNet at three different buffer sizes (5K exemplars=0.75GB, 10K exemplars=1.51GB, and 20K exemplars=3.01GB). To make the experiment fair, we compared REMIND to these models at equivalent buffer sizes, i.e., 479665 compressed samples=0.75GB, 959665 compressed samples=1.51GB, and 1281167 compressed samples (full dataset)=2.01 GB. In Fig.~\ref{fig:imagenet-stream-memory}, we see that more memory generally results in better performance. Overall, REMIND has the best performance and is nearly unaffected by buffer size. A plot with incremental batch models is in supplemental materials (Fig.~\ref{fig:imagenet-batch-memory}), and follows the same trend: larger buffers yield better performance.

\subsubsection{Regularization Comparisons.}

\begin{wraptable}[16]{r}{0.5\textwidth}
\caption{$\Omega_{\textrm{all}}$ for regularization models averaged over 10 runs on CORe50 with and without Task Labels (TL). \label{table:regularization-results}}
\centering
\begin{tabular}{lcccc}
\toprule
& \multicolumn{2}{c}{\textsc{cls iid}} & \multicolumn{2}{c}{\textsc{cls inst}} \\
\cmidrule(r){2-5}
\textsc{Model} & \textsc{TL} & \textsc{No TL} & \textsc{TL} & \textsc{No TL} \\ 
\midrule
SI & 0.895 & 0.417 & 0.905 & 0.416 \\
EWC & 0.893 & 0.413 & 0.903 & 0.413 \\
MAS & 0.897 & 0.415 & 0.905 & 0.421 \\
RWALK & 0.903 & 0.410 & 0.912 & 0.417 \\
A-GEM & 0.925 & 0.417 & 0.916 & 0.421 \\
REMIND & \textbf{0.995} & \textbf{0.978} & \textbf{0.995} & \textbf{0.979} \\
\midrule
Offline & 1.000 & 1.000 & 1.000 & 1.000 \\
\bottomrule
\end{tabular}
\end{wraptable}

In Table~\ref{table:regularization-results}, we show the results of REMIND and regularization methods for combating catastrophic forgetting on CORe50 class orderings. These regularization methods constrain weight updates to remain close to their previous values and are trained on batches of data, where each batch resembles a \emph{task}. At test time, these models are provided with task labels, denoting which task an unseen sample came from. In our experiments, a \emph{task} consists of several classes, and providing task labels makes classification easier. We analyze performance when task labels are provided and when they are withheld. To evaluate REMIND and Offline with task labels, we mask off probabilities during test time for classes not included in the specific task.  Consistent with \cite{chaudhry2018riemannian,farquhar2018towards,kemker2018forgetting}, we find that regularization methods perform poorly when no task labels are provided. Regardless, REMIND outperforms all comparisons, both with and without task labels.

\section{Discussion \& Conclusion}
\label{sec:discussion}

We proposed REMIND, a brain-inspired replay-based approach to online learning in a streaming setting. REMIND  achieved state-of-the-art results for object classification. Unlike iCaRL, Unified, and BiC, REMIND can be applied to iid and instance ordered data streams without modification. Moreover, we showed that REMIND is general enough for tasks like VQA with almost no changes.

REMIND replays compressed (lossy) representations that it stores, rather than veridical (raw pixel) experience, which is more consistent with memory consolidation in the brain. REMIND's replay is more consistent with how replay occurs in the brain during waking hours. Replay also occurs in the brain during slow wave sleep~\cite{barnes2014slow,ji2007coordinated}, and it would be interesting to explore how to effectively create a variant that utilizes sleep/wake cycles for replay. This could be especially beneficial for a deployed agent that is primarily engaged in online learning during certain hours, and is engaged in offline consolidation in other hours.

Several algorithmic improvements could be made to REMIND. We initialized REMIND's quantization model during the base initialization phase. For deployed, on-device learning this could instead be done by pre-training the codebook on a large dataset, or it could be initialized with large amounts of unlabeled data, potentially leading to improved representations. Another potential improvement is using selective replay. REMIND randomly chooses replay instances with uniform probability. In early experiments, we also tried choosing replay samples based on distance from current example, number of times a sample has been replayed, and the time since it was last replayed. While none performed better than uniform selection, we believe that selective replay still holds the potential to lead to better generalization with less computation. Because several comparison models used ResNet-18, we also used ResNet-18 for image classification so that we could compare against these models directly. The ResNet-18 layer used for quantization was chosen to ensure REMIND's memory efficiency, but co-designing the CNN architecture with REMIND could lead to considerably better results. Using less memory, REMIND stores far more compressed representations than competitors. For updating the replay buffer, we used random replacement, which worked well in \cite{chaudhry2018riemannian,wu2019large}. We tried a queue and a distance-based strategy, but both performed nearly equivalent to random selection with higher computational costs. Furthermore, future variants of REMIND could incorporate mechanisms similar to \cite{parisi2018lifelong} to explicitly account for the temporal nature of incoming data. To demonstrate REMIND's versatility, we pioneered streaming VQA and established strong baselines. It would be interesting to extend this to streaming chart question answering~\cite{kafle2018dvqa,kafle2020answering,Kahou2017FigureQAAA}, object detection, visual query detection~\cite{acharya2019vqd}, and other problems in vision and language~\cite{kafle2019challenges}.

\paragraph{Acknowledgements.}
This work was supported in part by the DARPA/MTO Lifelong Learning Machines program [W911NF-18-2-0263], AFOSR grant [FA9550-18-1-0121], NSF award \#1909696, and a gift from Adobe Research. We thank NVIDIA for the GPU donation. The views and conclusions contained herein are those of the authors and should not be interpreted as representing the official policies or endorsements of any sponsor. We thank Michael Mozer, Ryne Roady, and Zhongchao Qian for feedback on early drafts of this paper.

\clearpage
%
%
\bibliographystyle{splncs04}
\bibliography{egbib}

\clearpage
\begin{center}
    {\Large Supplemental Material \normalsize}
\end{center}
\beginsupplement

\section{Parameter Settings}

\begin{table}
\caption{Training parameter settings for REMIND and Offline models.}
\label{table:parameters}
\centering
\begin{tabular}{lccccc}
\toprule
\textsc{Parameters} & \textsc{ImageNet} & \textsc{CORe50} & \textsc{TDIUC} & \textsc{CLEVR}\\
\midrule
Optimizer & SGD & SGD & Adamax & Adamax  \\
Learning Rate & 0.1 & 0.01 & 2e-3 & 3e-4 \\
Momentum & 0.9 & 0.9 & - & - \\
Weight Decay & 1e-4 & 1e-4 & - & - \\
Streaming Batch Size & 51 & 21 & 51  & 51 \\
Offline Batch Size & 128 & 256 & 512 & 64 \\
Offline Epochs & 90 & 40 & 20 & 20 \\
Offline LR Decay & [30,60] & [15,30] & - & - \\
\bottomrule
\end{tabular}
\end{table}

We provide parameter settings for REMIND and the offline models in Table~\ref{table:parameters}. For the image classification experiments, we use the ResNet-18 implementation from the PyTorch Torchvision package. For the offline ImageNet model, we use standard data augmentation of random resized crops and random flips at 224$\times$224 pixels. We employ per-class learning rate decay for REMIND on ImageNet, using 0.1 as the starting learning rate and decaying it such that the learning rate becomes 0.001 after seeing all new samples for a class, at a step size of 100 new instances. For the $k$-means variant of REMIND, we use a codebook size of 10000 for ImageNet, and we found that increasing the codebook size yielded only marginal performance improvements. For CORe50, we do not use data augmentation with REMIND, as it harms performance. Unlike batch methods, REMIND learns one class at a time instance-by-instance.

To train REMIND on ImageNet in the incremental batch setting, we follow a paradigm similar to the incremental batch paradigm used by \cite{rebuffi2016icarl,wu2019large}. The base initialization stage for REMIND remains the same, where it trains offline on 100 classes and then subsequently trains the product quantizer and stores indices for previous examples in its memory buffer. We subject REMIND to the same buffer size during incremental batch learning as we do for streaming learning, which equates to 1.51  GB or compressed representations for 959665 examples. After base initialization, REMIND receives the next batch of 100 classes of data and mixes in all of the data from its replay buffer. It then loops over this data for 40 epochs, where the learning rate starts at 0.1 and is decayed by a factor of 10 at epochs 15 and 30. After looping over a batch, REMIND updates its replay buffer by storing new samples until it is full, and then randomly replacing samples from the class with the most examples. Consistent with our streaming experiments, the incremental batch version of REMIND uses random resized crops and mixup augmentation. 

For ExStream on the image classification experiments, we use 20 prototype vectors per class and the same parameters as the offline models. For SLDA on all experiments, we use shrinkage regularization of $10^{-4}$. Both ExStream and SLDA learn classes one at a time, instance-by-instance. For ImageNet, the parameters of iCaRL are kept the same as \cite{rebuffi2016icarl}. Similarly, the parameters for Unified and BiC on ImageNet are from \cite{hou2019unified} and \cite{wu2019large}, respectively. For the batch versions of CORe50 with iCaRL, Unified, and BiC, we train each batch for 60 epochs with a batch size of 64, weight decay of 1e-4, and a learning rate of 0.01 that we lower at epochs 20 and 40 by a factor of 5. For the streaming versions of iCaRL, Unified, and BiC, we set the number of epochs to 1 and the batch size to 51 and 21 for ImageNet and CORe50, respectively.

For MAC, we use the publicly available PyTorch implementation\\(\url{https://github.com/IBM/mi-prometheus}). For SAN, we use our own PyTorch implementation. For ExStream on TDIUC, we use an MLP with layer sizes [4096, 1024, 1480], lr = 2e-3, dropout with probability 0.5, Adamax optimizer, batch size of 512, and store 2500 exemplars per class. For ExStream on CLEVR, we use an MLP with layer sizes [3072, 1024, 28], lr = 2e-3, dropout with probability 0.5, Adamax optimizer, batch size of 512, and store 65 exemplars per class. For TDIUC and CLEVR, we chose the number of exemplars to consist of roughly 10\% of the dataset size.

\section{Where Should ResNet-18 be Quantized?}

\begin{figure}[t]
    \centering
    \begin{subfigure}
        \centering
        \includegraphics[width=0.45\linewidth]{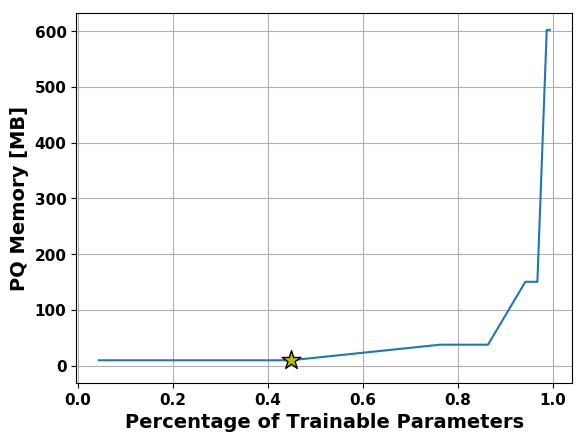}
    \end{subfigure}
    \begin{subfigure}
        \centering
        \includegraphics[width=0.45\linewidth]{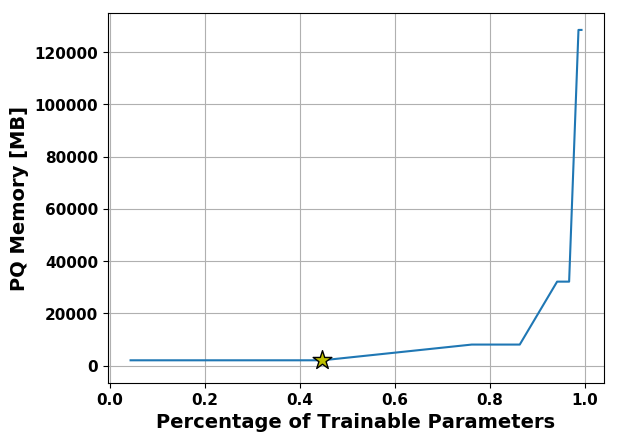}
    \end{subfigure}
    \caption{Auxiliary storage required to store quantized CNN features for the entire dataset as a function of the percentage of ResNet-18 parameters used in the top of the CNN, $F(\cdot )$, which are updated during streaming learning in REMIND. Storage requirements are shown for CORe50 (left) and ImageNet (right). The star denotes parameters used for our main experiments.   \label{fig:resnet-memory}}
\end{figure}

Following others, we used ResNet-18 for our incremental learning image classification experiments. This constrained the layers we could choose for quantization. If we quantized earlier in the network, the spatial dimensions of the feature tensor would be too large, resulting in much greater auxiliary storage requirements (see Fig.~\ref{fig:resnet-memory}). For example, in our ImageNet experiments, if we chose layer 3 of ResNet-18 for quantization, it would require 129  GB to store a representation of the entire training dataset; in contrast, the layer we used in our main experiments would require only 2  GB to store the entire training set. It is also a more biologically sensible layer to choose based on the connectivity of the hippocampus to visual processing areas.

If the architecture of ResNet-18 was altered to decrease the spatial dimensions earlier in the network, with a corresponding increase in the feature dimensions, this would allow us to quantize earlier in the network. However, this would prevent us from comparing directly to prior work and may require a considerable amount of architectural search to find a good compromise.

\section{Additional Image Classification Experiments}

\subsection{Buffer Size Comparisons}

\begin{wrapfigure}[15]{r}{0.45\textwidth}
 \centering
      \includegraphics[width=0.99\linewidth]{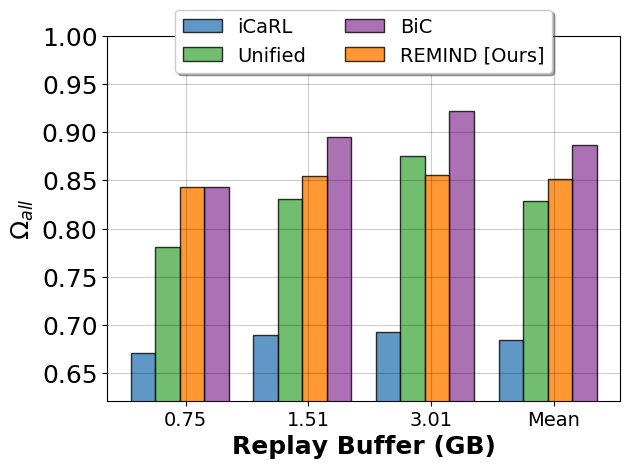}
  \caption{Performance as a function of buffer size for various batch comparison models on ImageNet. \label{fig:imagenet-batch-memory} 
  }
\end{wrapfigure}

Since REMIND and several other comparison models use replay as their main mechanism for mitigating forgetting, we were interested in examining how changes to the replay buffer size affected model performance on both ImageNet and CORe50. In Fig.~\ref{fig:imagenet-batch-memory}, we compare the performance of the incremental batch versions of iCaRL, Unified, and BiC on ImageNet at buffer sizes of 5K exemplars = 0.75 GB, 10K exemplars = 1.51 GB, and 20K exemplars = 3.01 GB, which are equivalent to REMIND storing 479665 compressed samples = 0.75 GB, 959665 compressed samples = 1.51 GB, and 1281167 compressed samples (full dataset) = 2.01 GB respectively. Note that this plot shows the performance of iCaRL, Unified, and BiC in the batch setting, but shows REMIND in the streaming setting, which is consistent with our main experiments.

These results demonstrate that REMIND and BiC are the top performers when a memory buffer of 0.75 GB is used, but BiC is the top performer when a memory buffer of 1.51 GB or 3.01 GB is used. However, REMIND rivals BiC's performance at both of these larger buffer sizes, only underperforming by 4\% and 6.6\% at 1.51 GB and 3.01 GB, respectively. It should be noted that BiC requires nearly 65 hours to train in incremental batch mode on ImageNet with a buffer size of 1.51 GB, whereas REMIND requires less than 12 hours with the same buffer size. Additionally, REMIND's performance is less dependent on the size of the buffer than BiC. That is, the difference between REMIND's performance at 0.75 GB and 3.01 GB is only 1.3\% in terms of $\Omega_{\mathrm{all}}$, whereas the difference between BiC's performance is 7.9\%, indicating that BiC is highly sensitive to the amount of storage allotted for replay. Additionally, while comparison models require 3.01 GB for the largest buffer size, REMIND's buffer size never exceeds 2.01 GB. Regardless, REMIND still achieves remarkable performance and rivals the state-of-the-art BiC model, even in the incremental batch setting.

\begin{figure}[t]
    \centering
    \begin{subfigure}
        \centering
        \includegraphics[width=0.45\linewidth]{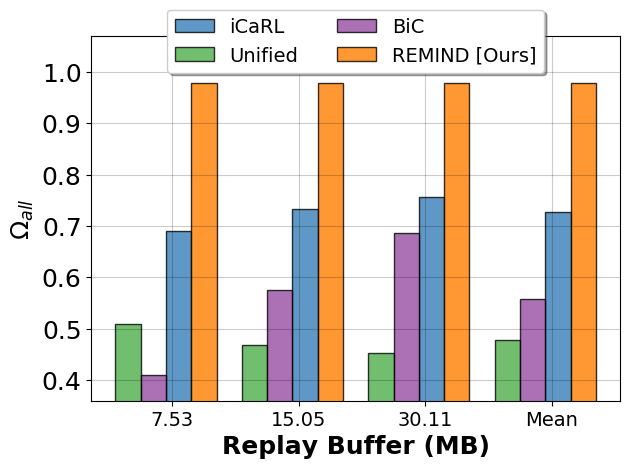}
    \end{subfigure}
    \begin{subfigure}
        \centering
        \includegraphics[width=0.45\linewidth]{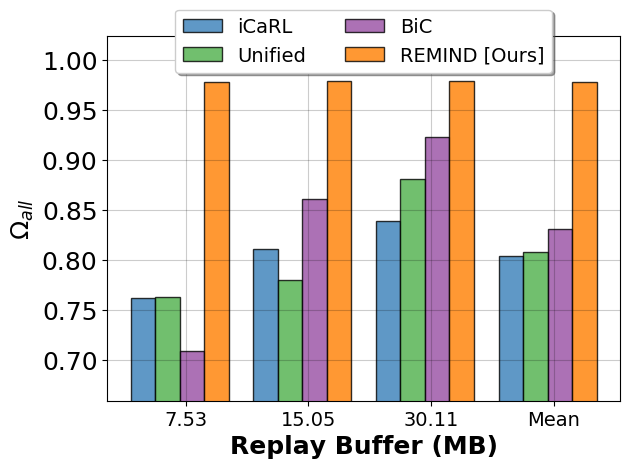}
    \end{subfigure}
    \caption{Performance as a function of buffer size for various streaming (left) and batch (right) comparison models on CORe50. Each bar is the average over 10 permutations. \label{fig:core50-memory}}
\end{figure}

In Fig.~\ref{fig:core50-memory}, we study the performance of the same models in both streaming and incremental batch mode on the CORe50 dataset. We study the performance of iCaRL, Unified, and BiC with buffer sizes of 50 exemplars = 7.53 MB, 100 exemplars = 15.05 MB, and 200 exemplars = 30.11 MB, which are equivalent to storing 4465 compressed samples = 7.53 MB and 6000 compressed samples (full dataset) = 9.93 MB respectively for REMIND. Our model is run in the streaming paradigm for both plots and outperforms all comparison models, regardless of the training paradigm, across all buffer sizes. This is remarkable since REMIND uses only \sfrac{1}{3} the amount of memory as compared to comparison models at 200 exemplars. Moreover, all of these comparison models use large amounts of additional memory to cache the information needed for distillation before learning a batch, which REMIND does not require.

\subsection{Updating Only $\theta_F$}
\label{sec:updating-only-F}
Since REMIND only updates $\theta_F$, it begs the question: is REMIND's superior performance a result of keeping $\theta_G$ fixed during incremental training? To answer this, we explore how other models perform when only $\theta_F$ is updated. On ImageNet, iCaRL, Unified, and BiC experience an absolute drop in $\Omega_{\textrm{all}}$ performance by 10.6\%, 2.7\%, and 2.8\%, respectively when only $\theta_F$ is plastic. This performance degradation indicates that this architectural choice actually harms competitors and does not provide REMIND with an unfair advantage.

\subsection{Changing $F(\cdot)$ and $G(\cdot)$} 

\begin{figure}[t]
\centering
\begin{tabular}
{c >{\centering\arraybackslash}m{2.2in} >{\centering\arraybackslash}m{2.2in} c}
	&
	\includegraphics[scale=0.35]{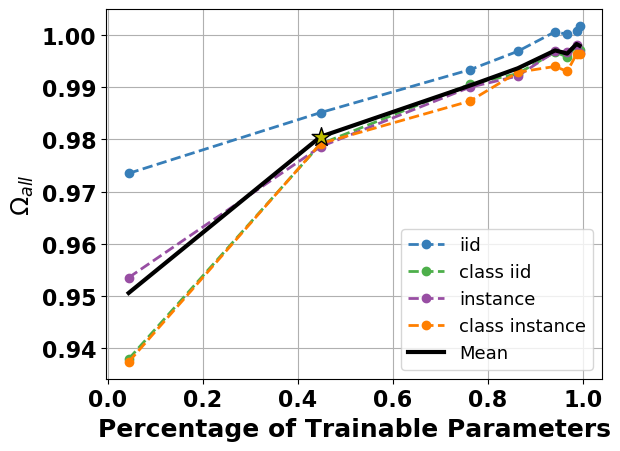} &
	\includegraphics[scale=0.35]{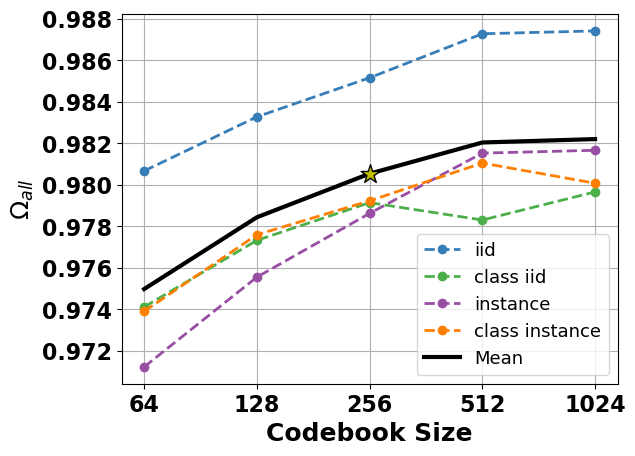} & \\
    &
	\includegraphics[scale=0.35]{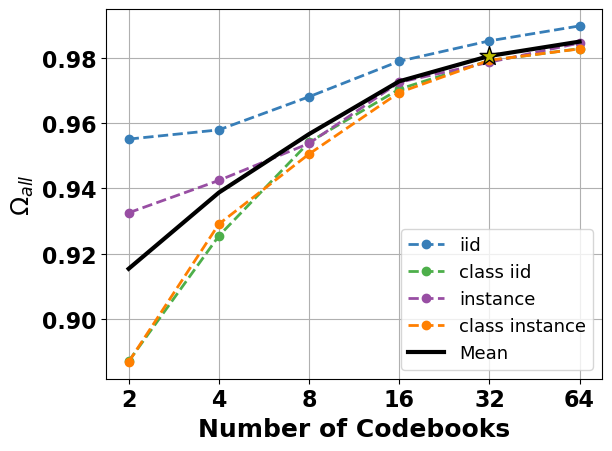} &
	\includegraphics[scale=0.35]{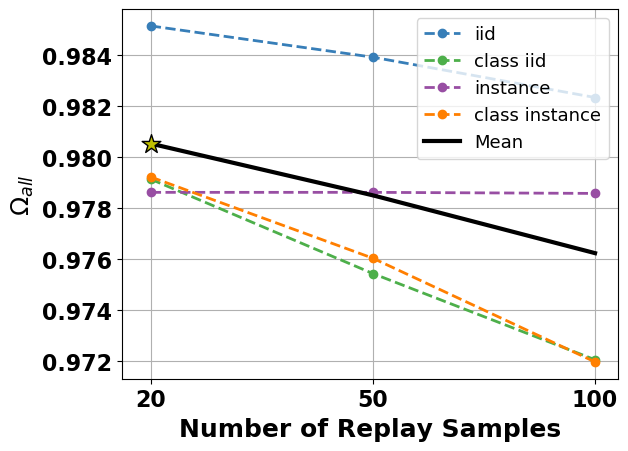} & \\
\end{tabular}
\caption{Additional experiments with REMIND on CORe50. From left to right, top to bottom, performance as a function of: 1) trainable parameters, 2) codebook size, 3) number of codebooks, and 4) number of replay samples ($r$). The values used for our main experiments are denoted with a yellow star and each dashed line is the average of 10 runs.}
\label{fig:additional-experiments-cls}
\end{figure}

One of the novelties of REMIND is the use of mid-level CNN features for training $\theta_F$. However, choosing where to extract features to train the PQ is an open question. In Fig,~\ref{fig:additional-experiments-cls}, we find that adding more trainable layers to $\theta_F$ improves accuracy on CORe50, but it has diminishing returns and there is a greater memory burden since features earlier in the network have larger spatial dimensions.

\subsection{Varying PQ Settings} 

REMIND's performance is dependent on the quality of tensor reconstructions used for training $F(\cdot)$. Since we use PQ to reconstruct samples from the replay buffer for REMIND, the performance is dependent on: 1) the number of codebooks used and 2) the size of the codebooks. We study performance on CORe50 as a function of the number of codebooks and codebook size in Fig.~\ref{fig:additional-experiments-cls}. We find that the performance improves as the number of codebooks and codebook size increase. However, memory efficiency decreases when these values are increased, so, we choose the number of codebooks to be 32 and codebook size to be 256 for our main experiments, making a trade-off between accuracy and memory efficiency. Since REMIND's performance is nearly unaffected by storing only 4465 samples compared to 6000, i.e., the entire CORe50 dataset, (see Fig.~\ref{fig:core50-memory}), we store the entire training set in the replay buffer for these additional studies.

\subsection{Altering Replay}
In our main experiments, each replay set contained 20 and 50 reconstructed samples for CORe50 and ImageNet, respectively. In Fig.~\ref{fig:additional-experiments-cls}, we study performance on CORe50 as a function of the number of replay samples. We found that performance degrades on CORe50 when we use more than 20 samples for replay. We hypothesize that since CORe50 has fewer samples, larger replay sizes cause overfitting, thereby degrading the performance. However, performance increases by 0.6\% for ImageNet (in terms of $\Omega_{\mathrm{all}}$), when the number of replay samples is increased from 20 to 50, which is the reason for using 50 samples in our main ImageNet experiments. Similar to the study of various PQ settings with REMIND on CORe50, we again store the entire training set in the replay buffer for this study on CORe50 due to the negligible performance difference (see Fig.~\ref{fig:core50-memory}).

\subsection{Average Accuracy for ImageNet and CORe50}

\begin{table}[t]
\caption{Average accuracy ($\mu_{\mathrm{all}}$) results for each dataset and ordering. For CORe50, we report the average over 10 runs. The best \emph{streaming} model for each dataset and ordering is highlighted in \textbf{bold}.
}
\label{tab:averaged-results}
\centering
\begin{tabular}{clccccc}
\toprule
& &  \textsc{\textbf{ImageNet}} & \multicolumn{4}{c}{\textsc{\textbf{CORe50}}} \\ 
 \cmidrule(r){3-3} \cmidrule(r){4-7}
\textsc{Model Type} & \textsc{Model} & \textsc{cls iid} & \textsc{iid} & \textsc{cls iid} & \textsc{inst} & \textsc{cls inst} \\ 
\midrule
\multirow{7}{*}{Streaming}
& Fine-Tune ($\theta_F$) & 26.80 & 88.72 & 11.95 & 76.27 & 11.95 \\
& ExStream & 52.65 & 87.97 & 48.01 & 83.72 & 46.91 \\
& SLDA  & 69.28 & 90.16 & 53.87 & 86.52 & 53.99 \\
& iCaRL & 28.61 & - & 37.88 & - & 35.46 \\
& Unified & 56.77 & - & 23.18 & - & 24.00 \\
& BiC & 40.64 & - & 16.08 & - & 16.68 \\
& REMIND & \textbf{78.68} & \textbf{91.00} & \textbf{55.35} & \textbf{88.08} & \textbf{55.42} \\
\midrule
\multirow{3}{*}{Incremental Batch}
& iCaRL & 63.59 & - & 41.94 & - & 42.10 \\
& Unified & 76.56 & - & 40.03 & - & 41.19 \\
& BiC & 82.38 & - & 35.08 & - & 39.24 \\
& REMIND & 80.55 & - & - & - & - \\
\midrule
\multirow{2}{*}{Upper Bounds}
& Offline ($\theta_F$) & 85.52 & 91.32 & 55.80 & 88.56 & 55.88 \\
& Offline & 91.95 & 92.35 & 56.99 & 89.93 & 56.94 \\
\bottomrule
\end{tabular}
\end{table}
\begin{table}[ht!]
\caption{Average accuracy ($\mu_{\mathrm{all}}$) results for CORe50 with their associated standard deviations over 10 runs with  different permutations of the data. The streaming models are at the top of the table, while the upper bounds are at the bottom. The best model for each ordering is highlighted in \textbf{bold}.
}
\label{tab:averaged-results-core50}
\centering
\begin{tabular}{lcccc}
\toprule
\textsc{Model} & \textsc{iid} & \textsc{cls iid} & \textsc{inst} & \textsc{cls inst} \\ 
\midrule
Fine-Tune ($\theta_F$) & 88.72$\pm$1.57 & 11.95$\pm$0.02 & 76.27$\pm$4.44 & 11.95$\pm$0.03 \\
ExStream & 87.97$\pm$0.83 & 48.01$\pm$2.17 & 83.72$\pm$1.78 & 46.91$\pm$2.35 \\
SLDA & 90.16$\pm$0.63 & 53.87$\pm$0.79 & 86.52$\pm$1.12 & 53.99$\pm$0.82 \\
iCaRL & - & 37.88$\pm$3.41 & - & 35.46$\pm$2.89 \\
Unified & - & 23.18$\pm$5.47 & - & 24.00$\pm$5.69 \\
BiC & - & 16.08$\pm$1.93 & - & 16.68$\pm$2.00 \\
REMIND & \textbf{91.00$\pm$0.58} & \textbf{55.35$\pm$0.95} & \textbf{88.08$\pm$1.33} & \textbf{55.42$\pm$0.86} \\
\midrule
Offline ($\theta_F$) & 91.32$\pm$0.42 & 55.80$\pm$0.61 & 88.56$\pm$1.04 & 55.88$\pm$0.60 \\
Offline & 92.35$\pm$0.40 & 56.99$\pm$0.48 & 89.93$\pm$0.78 & 56.94$\pm$0.46 \\
\bottomrule
\end{tabular}
\end{table}

In the main paper, we present $\Omega_{\textrm{all}}$, which makes it easy to compare across datasets, orderings, and paradigms. However, it can hide the raw performance of the models. Following others~\cite{castro2018end,hou2019unified,rebuffi2016icarl,wu2019large}, we provide the average accuracy metric over all testing intervals, i.e., 
\begin{equation}
\mu_{\mathrm{all}} = \frac{1}{T} \sum_{t=1}^T \alpha_{t} \enspace ,
\end{equation}
where $T$ is the total number of testing events and $\alpha_{t}$ is the accuracy of the model for test $t$. We provide $\mu_{\mathrm{all}}$ results in Table~\ref{tab:averaged-results}, which shows the top-5 accuracy for ImageNet and top-1 accuracy for CORe50. When using these metrics, REMIND is still the top streaming performer and competitive in the incremental batch setting on ImageNet. CORe50 results for the class orderings are lower because we test on all test data at every interval, which includes classes that are yet to be seen. This leads to low accuracies for the unseen classes, which affects $\mu_{\mathrm{all}}$.

On CORe50 we also report the average accuracy and associated standard deviation values over 10 runs with different permutations of the dataset in Table~\ref{tab:averaged-results-core50}. Overall, the iid and instance orderings yielded the highest model performances, making them easiest, while the class orderings resulted in much worse performance, making them hardest. REMIND’s results are statistically significantly different from each of the comparison models for all four data orderings according to a Student’s t-test at a 99\% confidence interval.

\section{Additional VQA Experiments}

\subsection{REMIND Performance for Various Buffer Sizes}

\begin{table}[h]
\caption{$\Omega_{\mathrm{all}}$ results for REMIND with various buffer sizes on streaming VQA.  \label{tab:vqa-ablations}}
\centering
\resizebox{0.5\textwidth}{!}{%
\begin{tabular}{lcccc}
\toprule
& \multicolumn{2}{c}{\textsc{TDIUC}} & \multicolumn{2}{c}{\textsc{CLEVR}} \\
\cmidrule(r){2-3} \cmidrule(r){4-5}
\textsc{Buf. Size} & \textsc{iid} & \textsc{q-type} & \textsc{iid} & \textsc{q-type} \\ 
\midrule
25\% & 0.914 & \textbf{0.936} & \textbf{0.724} & 0.960 \\
50\% & 0.917 & 0.919 & 0.720 & 0.979 \\
75\% & \textbf{0.919} & 0.914 & 0.722 & 0.984 \\
100\% & 0.914 & 0.931 & 0.723 & \textbf{0.985} \\
\midrule
Offline & 1.000 & 1.000 & 1.000 & 1.000 \\
\bottomrule
\end{tabular}
}
\end{table}

In Table~\ref{tab:vqa-ablations}, we provide additional results for REMIND on TDIUC and CLEVR with buffer sizes that consist of 25\%, 50\%, 75\%, and 100\% of the samples from the entire training set. Overall, we see that REMIND performs remarkably well with a limited buffer size. For example, the model trained with only a 25\% buffer size rivals, and in some cases outperforms, the model with a 100\% buffer size.

\subsection{Learning Curves and Qualitative Examples}

\begin{figure}[t]
\centering
\begin{tabular}
{c >{\centering\arraybackslash}m{2.2in} >{\centering\arraybackslash}m{2.2in} c}
	&
	\includegraphics[scale=0.35]{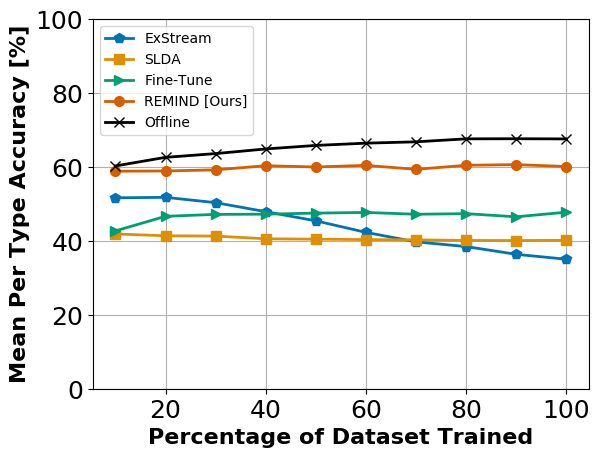} &
	\includegraphics[scale=0.35]{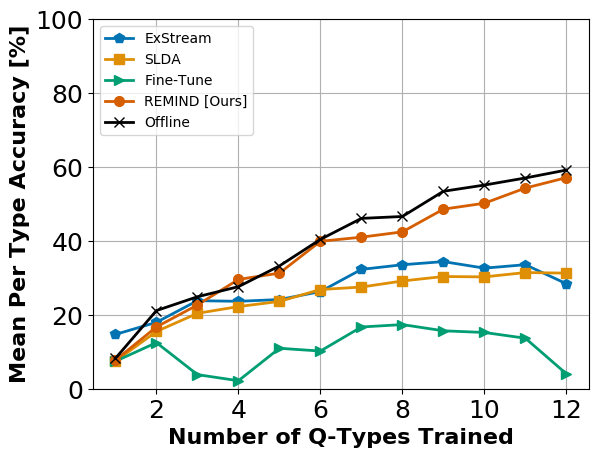} & \\
    &
	\includegraphics[scale=0.35]{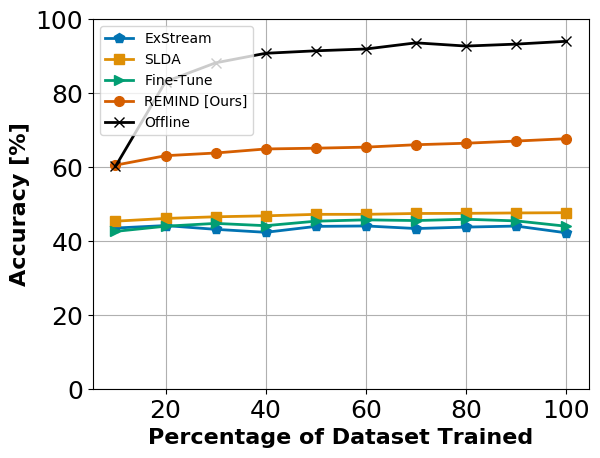} &
	\includegraphics[scale=0.35]{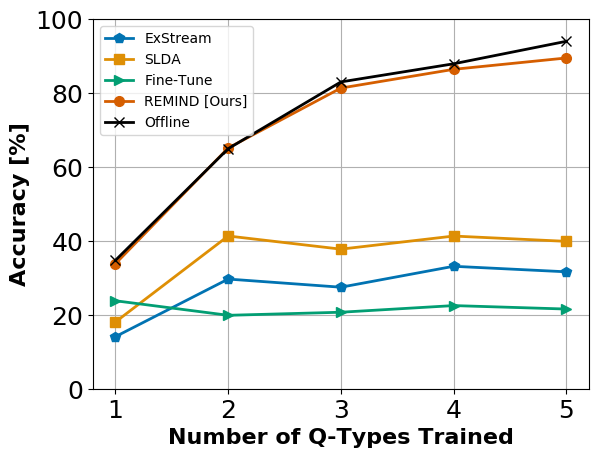} & \\
\end{tabular}
    \caption{Learning curves for each ordering of the TDIUC (top row) and CLEVR (bottom row) datasets. We provide curves from the REMIND model trained with 50\% buffer size.}
    \label{fig:main-results-vqa}
\end{figure}

\begin{figure}[t]
    \centering
    \begin{subfigure}
        \centering
        \includegraphics[width=0.8\textwidth]{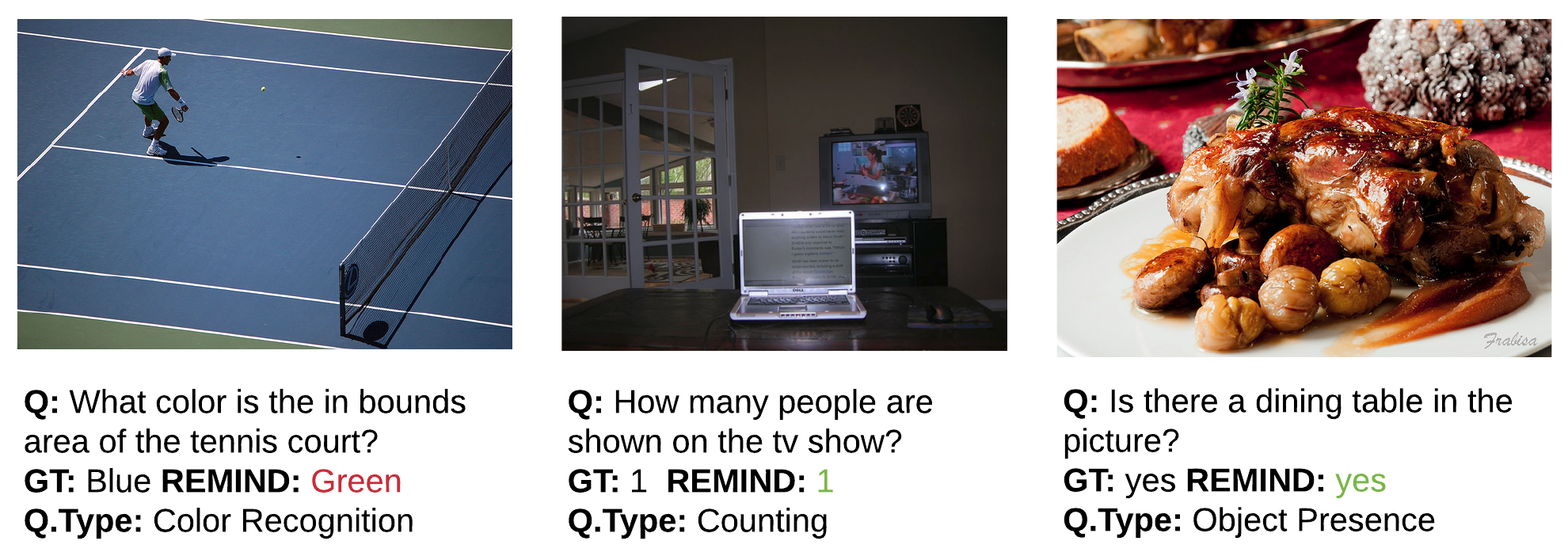}
    \end{subfigure} 
    \begin{subfigure}
        \centering
        \includegraphics[width=0.8\textwidth]{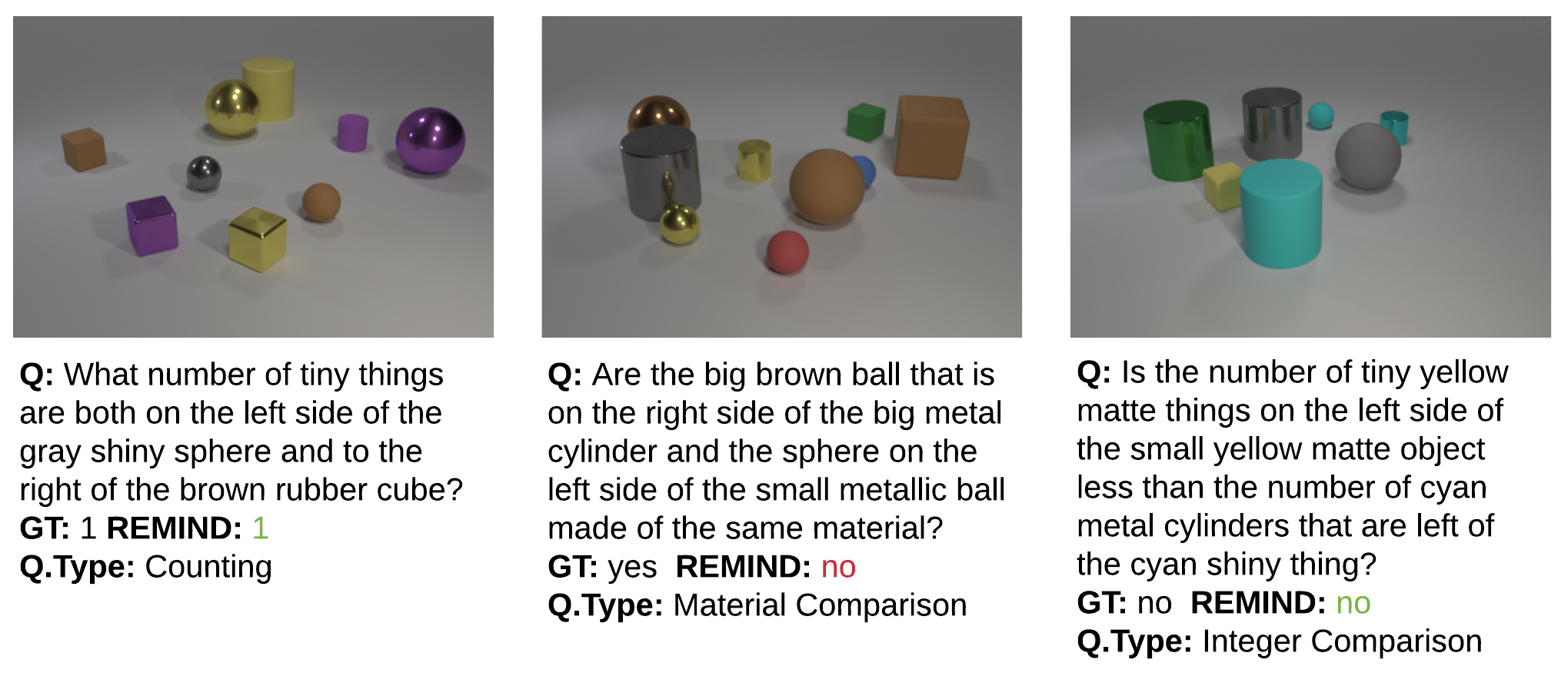}
    \end{subfigure} 
    \caption{Qualitative VQA examples on the TDIUC (top row) and CLEVR (bottom row) datasets. We provide examples from the REMIND model trained with 50\% buffer size on the q-type ordering for both datasets.}
    \label{fig:qualitative-vqa}
\end{figure}

We provide learning curves for each of the main VQA experiments in Fig.~\ref{fig:main-results-vqa} and qualitative examples in Fig.~\ref{fig:qualitative-vqa}. REMIND's learning curve closely follows the offline curve for the q-type ordering of both the TDIUC and CLEVR datasets. This indicates that our model is able to learn new q-types without forgetting old q-types. For the iid ordering of TDIUC, the accuracy remains more or less constant after the first increment and for the iid ordering of CLEVR, the accuracy increases at a slower rate than the offline model. We believe that training with samples ordered by q-type may have acted as a natural curriculum for REMIND, providing more benefits to the VQA model.

\end{document}